\documentclass{article}

\PassOptionsToPackage{numbers, compress}{natbib}
\pdfoutput=1

\usepackage[nonatbib, preprint]{neurips_2025}

\usepackage[utf8]{inputenc} 
\usepackage[T1]{fontenc}    
\usepackage{hyperref}       
\usepackage{url}            
\usepackage{booktabs}       
\usepackage{amsfonts}       
\usepackage{nicefrac}       
\usepackage{microtype}      
\usepackage{xcolor}         
\usepackage{amsmath}
\usepackage{bm}
\usepackage{mathtools}
\usepackage{algorithm}

\usepackage{algpseudocode}

\usepackage{multirow}
\usepackage{colortbl}
\usepackage{wrapfig}
\DeclareMathOperator*{\argmax}{arg\,max}
\DeclareMathOperator*{\argmin}{arg\,min}

\title{Towards more transferable adversarial attack in black-box manner}

\author{Chun Tong Lei$^1$, Zhongliang Guo$^2$, Hon Chung Lee$^1$, Minh Quoc Duong$^1$, Chun Pong Lau$^1$\\
{\normalsize$^1$City University of Hong Kong} \quad {\normalsize$^2$University of St Andrews}\\
{\small\{ctlei2, honclee8, mduong2, cplau27\}@cityu.edu.hk,}
{\small zg34@st-andrews.ac.uk
    }}

\begin{document}

\maketitle
\begin{abstract}
Adversarial attacks have become a well-explored domain, frequently serving as evaluation baselines for model robustness.
Among these, black-box attacks based on transferability have received significant attention due to their practical applicability in real-world scenarios.
Traditional black-box methods have generally focused on improving the optimization framework (e.g., utilizing momentum in MI-FGSM or other variants like NI-FGSM) to enhance transferability, rather than examining the dependency on surrogate white-box model architectures.
Recent state-of-the-art approach DiffPGD has demonstrated enhanced transferability by employing diffusion-based adversarial purification models for adaptive attacks.
The inductive bias of diffusion-based adversarial purification aligns naturally with the adversarial attack process, where both involving noise addition, reducing dependency on surrogate white-box model selection.
However, the denoising process of diffusion models incurs substantial computational costs through chain rule derivation, manifested in excessive VRAM consumption and extended runtime.
This progression prompts us to question whether introducing diffusion models is necessary.
We hypothesize that a model sharing similar inductive bias to diffusion-based adversarial purification, combined with an appropriate loss function, could achieve comparable or superior transferability while dramatically reducing computational overhead.
In this paper, we propose a novel loss function coupled with a unique surrogate model to validate our hypothesis.
Our approach leverages the score of the time-dependent classifier from classifier-guided diffusion models, effectively incorporating natural data distribution knowledge into the adversarial optimization process.
Experimental results demonstrate significantly improved transferability across diverse model architectures while maintaining robustness against diffusion-based defenses.

\end{abstract}
\section{Introduction}
Adversarial attacks represent a significant challenge to the reliability of deep learning systems~\cite{szegedy2013intriguing,guo2024grey,zhao2025a}. By introducing carefully crafted, imperceptible perturbations to input data, these attacks can cause state-of-the-art neural networks to produce erroneous outputs with high confidence. The vulnerability of deep neural networks to such manipulations raises serious concerns about their deployment in security-critical applications~\cite{guo2024white,li2025threats,zhao2024weak}, from autonomous vehicles to medical diagnostics systems.

Among the various adversarial attack strategies, black-box attacks have emerged as particularly concerning due to their practical applicability. These attacks assume limited knowledge of the target model's architecture and parameters, reflecting real-world scenarios where attackers have restricted access to deployed systems.
Transfer-based black-box attacks, which generate adversarial examples on accessible surrogate models and apply them to target models, have proven remarkably effective despite their conceptual simplicity~\cite{dong2018boosting,guo2024artwork}.
The efficacy of transfer-based attacks is fundamentally tied to the concept of transferability, i.e., the ability of adversarial examples to fool multiple models beyond those they were specifically crafted to deceive. Traditional approaches mainly focus on improve the optimization framework for better transferability~\cite{dong2018boosting}, besides, ensemble methods leverage multiple models to achieve the same goal~\cite{feng2021meta}.
However, these approaches suffer from significant limitations: their performance remains highly dependent on the surrogate model, while the latter often requiring resource-intensive ensemble attacks to achieve reliable results.

\begin{figure*}[t]
    \centering
    \includegraphics[width=\linewidth]{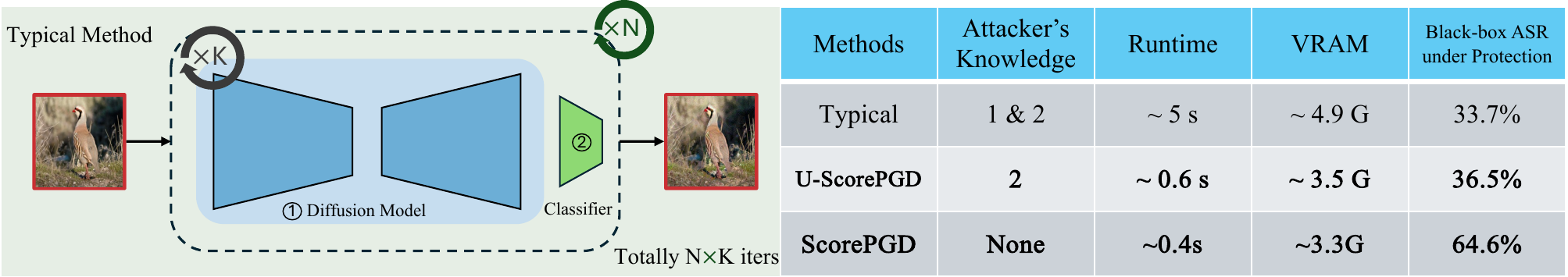}
    \includegraphics[width=\linewidth]{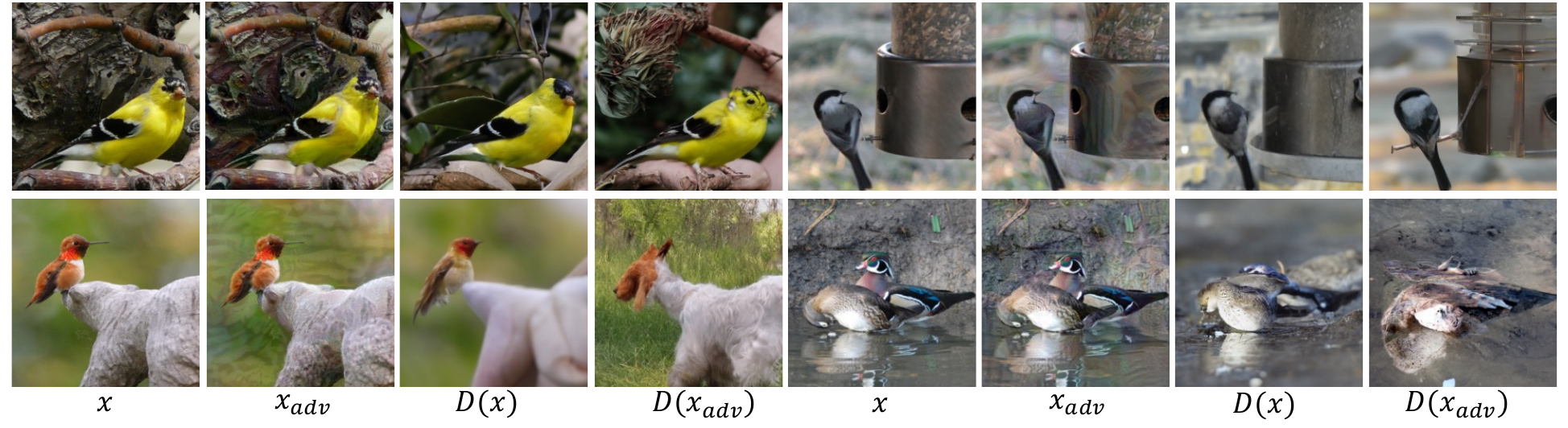}

    \caption{ The above part shows that our method uses less runtime and VRAM compared to the current SOTA method DiffPGD~\cite{xue2023diffusion}, meanwhile having more black-box transferability under protection by diffusion-based purification method. The result on black-box transferability is tested on ResNet101, more details can be found in the experiment section.
    The bottom part visualizes the capability of our proposed ScorePGD regarding disrupting image editing task. The setting for generating these images is $\gamma=8/255$ and $t=0.5T$ for the diffusion model $D$.}
    \label{fig:teaser}
\end{figure*}

Recent advances have attempted to overcome these limitations through innovative methodologies. DiffPGD~\cite{xue2023diffusion} represents the current state-of-the-art, employing diffusion-based adversarial purification models~\cite{nie2022diffusion} within the attack pipeline to significantly enhance transferability. By leveraging the inductive bias of diffusion models, which naturally aligns with the adversarial attack process through systematic noise addition and removal, DiffPGD reduces dependence on surrogate model selection and eliminates the need for ensemble attacks.

Despite its impressive performance, DiffPGD introduces substantial computational overhead. The denoising process inherent to diffusion models necessitates multiple applications of complex neural networks, resulting in excessive VRAM consumption and extended processing times.
This computational burden severely limits practical deployment, particularly in resource-constrained environments or large-scale robustness evaluation.

This limitation motivates our central research question:
\begin{align*}
    &\text{\textit{Can we develop an attack methodology that achieves comparable or superior transferability}}\\
    &\text{\textit{ to diffusion-based approaches while dramatically reducing computational requirements?}}
\end{align*}
We hypothesize that the critical factor enhancing transferability in DiffPGD is not the complete diffusion process itself, but rather the incorporation of noised data distribution knowledge into the adversarial optimization process.

Based on the above motivation, in this paper, we introduce a novel attack method that validates this hypothesis.
Our approach leverages the score of the time-dependent classifier from classifier-guided diffusion models~\cite{dhariwal2021diffusion}, effectively capturing the essential distribution information without requiring the complete diffusion machinery. By pairing this with a carefully designed loss function and surrogate model architecture, we achieve two significant advances:
\begin{itemize}
    
    \item Enhanced transferability across diverse model architectures, including both standard models and those protected by diffusion-based defenses.
    \item Order-of-magnitude reduction in computational requirements compared to DiffPGD.
\end{itemize}

Our method is much faster than DiffPGD, meanwhile successfully attack the diffusion model and classifier without any knowledge from the target model (as shown in Figure~\ref{fig:teaser}). 

Our extensive experiments demonstrate that the proposed method not only matches but frequently surpasses the performance of state-of-the-art approaches while drastically reducing resource demands. These results suggest that efficient incorporation of noised data distribution knowledge, rather than the full access to this information, may be the critical factor in developing highly transferable adversarial attacks.

The significance of our work extends beyond the immediate performance improvements. By decoupling transferability enhancement from computational intensity, we enable the deployment of advanced adversarial attacks in previously prohibitive contexts. Additionally, our findings provide valuable insights into the fundamental mechanisms underlying transferability in adversarial examples, potentially informing the development of more robust defense strategies.

\section{Related Works}

\begin{figure}[t]
    \centering
    \includegraphics[width=\linewidth]{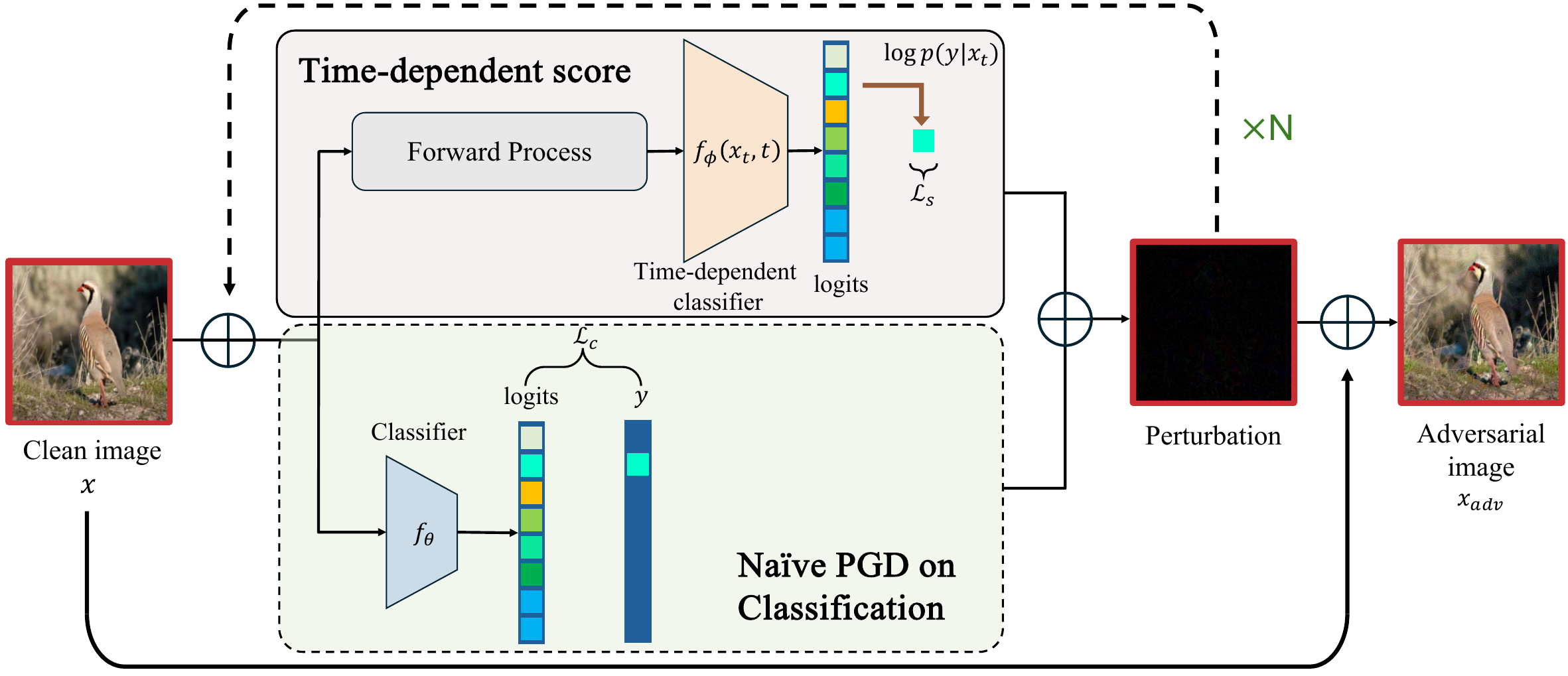}

    \caption{The illustration of our method. The $f_\phi(x_t,t )$ is a classifier trained on noised image, with the noise scale up by timestep $t$. We calculate the cross entropy loss $\mathcal{L}_c$, which is optional, and the log-likelihood of the ground truth label simultaneously. The variant with calculating $\mathcal{L}_c$ is U-ScorePGD, the variant without that is ScorePGD. Then we are trying to iteratively maximizing the cross entropy and minimizing the log-likelihood in the optimization process.}
    \label{fig:pipeline}
\end{figure}

\paragraph{Adversarial Attack}
Neural networks' vulnerability to subtle perturbations was first identified by Szegedy et al.\cite{szegedy2013intriguing}, revealing a fundamental weakness where minor input modifications can cause incorrect classifications. Building on this discovery, Goodfellow et al.\cite{goodfellow2015explaining} introduced the concept of adversarial examples and proposed the Fast Gradient Sign Method (FGSM). Subsequently, more sophisticated attacks emerged, including Projected Gradient Descent (PGD)\cite{madry2018towards}, which iteratively maximizes the model's loss function within an $\epsilon$-ball constraint, and momentum-based approaches
\cite{dong2018boosting} that enhance attack stability and effectiveness.
\paragraph{White-Box and Black-Box Attacks}
Adversarial attacks are typically categorized as white-box or black-box. White-box attacks~\cite{madry2018towards, goodfellow2015explaining} leverage complete access to the target model's architecture and parameters, enabling highly effective adversarial examples through gradient backpropagation. Black-box attacks operate with limited model access, making them more relevant to real-world scenarios. These approaches either use query-based methods~\cite{brendel2017decision, chen2017zoo} or exploit cross-model transferability~\cite{papernot2016transferability, chen2024rethinking, yao2025understanding, chen2024diffusion}, where perturbations created using surrogate models successfully deceive target models.
\paragraph{Diffusion Model}
Diffusion models~\cite{sohl2015deep, ho2020denoising} have gained prominence for their generative capabilities, operating by adding Gaussian noise to images and then iteratively removing it using U-Net structures. Control methods for diffusion models include classifier-guided~\cite{dhariwal2021diffusion} and classifier-free guided approaches~\cite{ho2022classifier, rombach2022high}. In the adversarial context, DiffPure~\cite{nie2022diffusion,lei2025instant} leverages diffusion processes to remove perturbations from adversarial samples. DiffPGD~\cite{xue2023diffusion} enhances attack transferability by incorporating diffusion processes into attack iterations, transforming out-of-distribution perturbations into in-distribution attacks. However, DiffPGD's back-propagation is computationally expensive due to diffusion model chain rule operations, requiring multiple steps even with DDIM~\cite{songdenoising} acceleration.

\section{Preliminaries}

\subsection{Diffusion Model} Denoising Diffusion Probabilistic Models (DDPM)~\cite{ho2020denoising} is the pioneer work of diffusion model family, which includes two processes: forward process and reverse process. Forward process first sample $\bm{x}_0\sim q(\mathcal{X})$ where $\mathcal{X}$ is a set of natural images (e.g., the
space of images $\mathbb{R}^{H \times W \times C}$) and $q$ is a probability measure defined on $\mathcal{X}$. Then gradually adds Gaussian noise to the image $\bm{x}_0$ to generate a sequence of noisy samples $\{\bm{x}_t\}^T_{t=1}
$ with a scaling schedule $\{\alpha_t\}^T_{t=1}$ and a noise schedule
$\{\sigma_t\}^T_{t=1}$ by 
\begin{equation}
    q(\bm{x}_t|\bm{x}_0)=\mathcal{N}(\bm{x}_t;\sqrt{\alpha_t}\bm{x}_0,\sigma^2_t\mathbf{I}),
\label{forward process}
\end{equation}

which is a diagonal Gaussian distribution. Since the signal-to-noise ratio $\text{SNR}(t) = \alpha_t^2/\sigma^2_t$ is always defined as a strictly monotonically decreasing function of the timestep $t$, the sample $\bm{x}_t$ becomes more noisy with increasing $t$. The scaling and noise schedules are predefined with the intent of making $\bm{x}_T$ an isotropic Gaussian distribution. Reverse process is defined to approximate $q(\mathcal{X})$ by gradually removing noise from the standard Gaussian distribution $p(\bm{x}_T) = \mathcal{N}(\bm{x}_T; \mathbf{0},\mathbf{I})$:
\begin{equation}
\begin{split}
    &p(\bm{x}_{0:T}) = p(\bm{x}_T)\prod_{t=1}^T
p_{\theta}(\bm{x}_{t-1} \vert \bm{x}_t),\\
&p_\theta(\bm{x}_{t-1}\vert \bm{x}_t)=\mathcal{N}(\bm{x}_{t-1}; \mu_\theta(\bm{x}_t,t),\tilde 
\sigma^2_t\mathbf{I}),
\label{reverse process}
\end{split} 
\end{equation}
where $\mu_\theta$ is usually parameterized by a noise prediction model $\epsilon_\theta(\bm{x}_t,t)$~\cite{ho2020denoising, kingma2021variational}:
\begin{equation}
    \bm{\mu}_\theta(\bm{x}_t,t) = \sqrt{\frac{\alpha_{t-1}}{\alpha_t}}\left(\bm{x}_t-\frac{\sigma_t^2-\frac{\alpha_t}{\alpha_{t-1}}\sigma^2_{t-1}}{\sigma_t}
\bm{\epsilon}_\theta(\bm{x}_t,t)\right).
\end{equation}
The reverse step $p_{\theta}(\bm{x}_{t-1} \vert \bm{x}_t)$ is trained by optimizing the diffusion loss $\mathbb{E}_{\epsilon,t}[\vert\vert \bm{\epsilon}_\theta(\bm{x}_t,t)- \bm{\epsilon}\vert\vert^2_2] $

\begin{figure}[t]
    \centering
    \includegraphics[width=\linewidth]{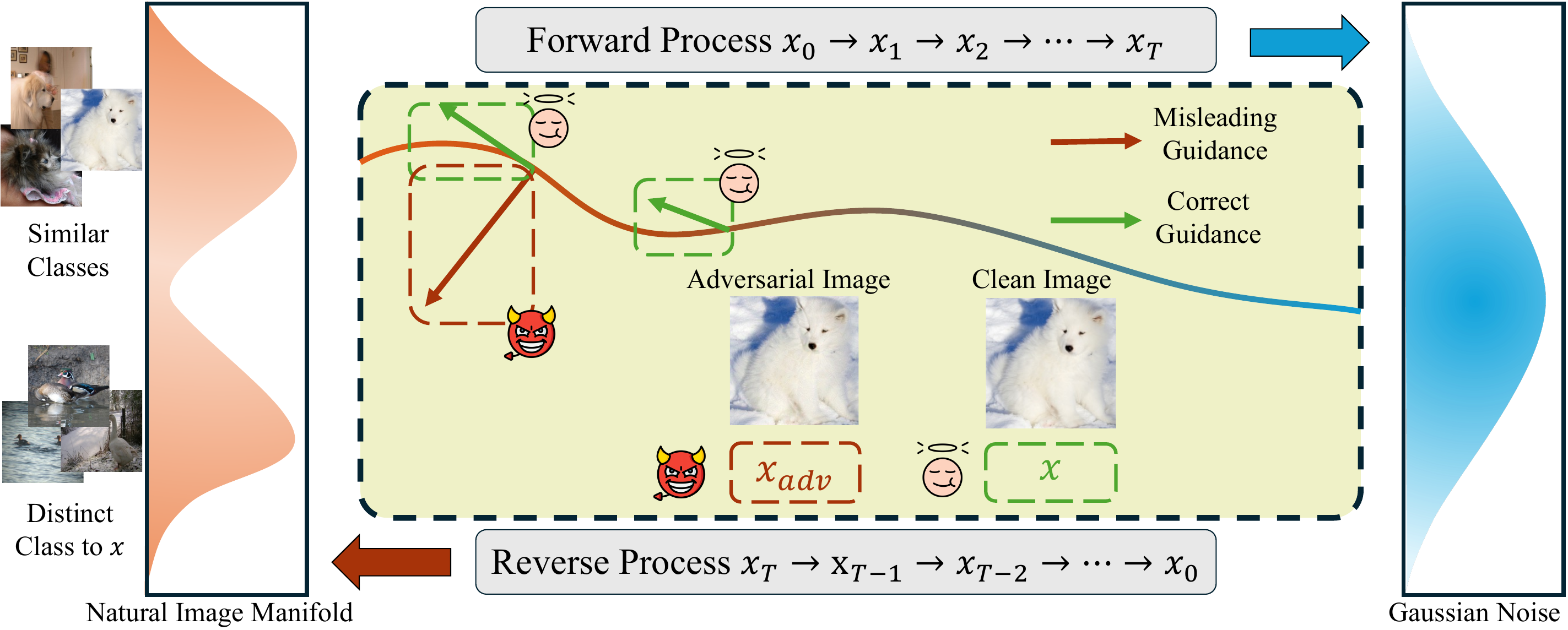}
    \caption{The illustration of the ScorePGD's objective. Our method aims to change the guidance direction of the reverse process of the diffusion model. The direction of guidance is the score of the time-dependent classifier, and hence our ScorePGD method will induce a distillation with wrong guidance information into the image by the perturbation, leading the diffusion editing purification deal with wrong class.}
    \label{fig:score target}
\end{figure}

\subsection{Conditional Reverse Noising Process} To have better controllability for generation, the pre-trained diffusion model can be conditioned using the gradients of a classifier \cite{sohl2015deep, song2020score}. In particular, we can use a classifier $p_\phi(y|\bm{x}_t,t)$ trained on noisy images $\bm{x}_t$, and then use score $\nabla_{\bm{x}_t} \log p_\phi(y\vert \bm{x}_t,t)$ to guide the diffusion sampling process towards an arbitrary class label $y$ by:
\begin{equation}
    p_{\theta,\phi}(\bm{x}_t|\bm{x}_{t+1},y) =Zp_\theta(\bm{x}_t|\bm{x}_{t+1})p_\phi(y|\bm{x}_t),
    \label{probaility bayes rule}
\end{equation}
where $Z$ is a normalizing constant. Then, the conditional reverse process is recursively sampling
\begin{equation}
\bm{x}_{t-1} \sim \mathcal{N}(\bm{\mu}_\theta(\bm{x}_t,t)+s\sigma_t\nabla_{\bm{x}_t} \log p_\phi(y\vert \bm{x}_t,t),\sigma_t\mathbf{I}),
\end{equation}
until $\bm{x}_0$. In addition to stochastic diffusion sampling process, deterministic sampling methods, such as Denoising Diffusion Implicit Models (DDIM)~\cite{songdenoising} can also combine with conditional sampling,
\begin{equation}
\bm{x}_{t-1}=\sqrt{\alpha_{t-1}}\left(\frac{\bm{x}_t-\sigma_t\hat{\bm{\epsilon}}(\bm{x}_t,t)}{\sqrt{\alpha_t}}\right)+\sigma_{t-1}\hat{\bm{\epsilon}}(\bm{x}_t,t),
\end{equation}
where $\hat{\bm{\epsilon}}(\bm{x}_t,t)\coloneq\bm{\epsilon}_\theta(\bm{x}_t,t)-\sigma_t\nabla_{\bm{x}_t} \log p_\phi(y\vert \bm{x}_t,t)$

\section{Method}

\subsection{Diffusion-based Purification}
\label{dbp}

Diffusion-based purification (DBP), a two-step adversarial purification method which leverages diffusion models. For any adversarial example $\bm{x}_{adv}=\bm{x}_0+\bm{\delta}$, where $\bm{\delta}$ is an adversarial noise added to natural image $\bm{x}_0$ at diffusion timestep $t=0$. First, diffuse it by the forward process in Equation~\ref{forward process} from $t=0$ to $t=t^\ast$, where $t^\ast$ is predefined:
\begin{equation}
    \bm{x}_{adv}^{t^\ast}=\sqrt{\alpha_t}\bm{x}_{adv}+\sigma_{t^\ast}\bm{\epsilon}, \quad\bm{\epsilon}\sim\mathcal{N}(\mathbf{0},\mathbf{I}).
    \label{noised adversarial sample}
\end{equation}
Second, we manipulate the reverse process in Equation~\ref{reverse process} from the timestep $t=t^\ast$ using the noised adversarial sample $\bm{x}_{adv}^{t^\ast}$,
given by Equation~\ref{noised adversarial sample}, to obtain the purified sample $\hat{\bm{x}}_0$ from iterating reverse step $p_{\theta}(\bm{x}_{t-1} \vert \bm{x}_t)$ in Equation~\ref{reverse process}. Nie et.\ al~\cite{nie2022diffusion} prove that the
KL divergence of $p_t$ and $q_t$, which are respective distributions of the noised natural image $\bm{x}_t$ and noised adversarial image $\bm{x}_{adv}^t$ for $\bm{x}\sim p(\mathcal{X})$, where $\mathcal{X}$ is a natural image set and $\bm{x}_{adv}\sim q(\mathcal{X}_{adv})$, $\mathcal{X}_{adv}$ is an adversarial image set, decreases monotonically from $t=0$ to $t=T$ through the forward process. Hence, we suggest that this is the reason making diffusion-based purification method robust to the adversarial samples.

\begin{algorithm}[t]
\caption{ScorePGD}
\label{algo:score}
\begin{algorithmic}
\Require Original image $\bm{x}$, time-dependent classifier $f_\phi(\bm{x}_t,t)$, predefined timestep $t$ for the time-dependent classifier, iteration $n$, stepsize $\eta$ and bound value $\gamma$,

\State $\bm{x}_{adv}=\bm{x}$
\For{$t=0,1,2,\dots,n-1$}

\State $\bm{\epsilon}\sim \mathcal{N}(\mathbf{0},\mathbf{I})$
\State $\bm{x}_{adv}^t=\sqrt{\alpha_t}\bm{x}_{adv}+\sigma_t\bm{\epsilon}$ \Comment{Sample $\bm{x}_{adv}^t$ for each PGD iteration by Equation~\ref{forward process}}
\State $\mathcal{L}_s = \log f_\phi(\bm{x}_{adv}^t,t) $ \Comment{Calculate score loss by Equation~\ref{scoreloss}}
\State $\mathcal{L}_s = - \mathcal{L}_s$ \Comment{Change the sign of the loss since minimizing}
\State $\bm{\delta}^{(i+1)}=\mathcal{P}_{\gamma}\left(\bm{\delta}^{(i)}+\eta \cdot \text{sign}\nabla_{\bm{\delta}^{(i)}}\mathcal{L}_s\right)$   \Comment{Update the value of $\delta$ by Equation~\ref{pgduqdate}}
\State $\bm{x}_{adv} = \bm{x}_{adv}+\bm{\delta}$ 
\EndFor
\State\Return $x_{adv}$
\end{algorithmic}
\end{algorithm}

\subsection{Score PGD}

Recent approaches like DiffPGD~\cite{xue2023diffusion} have demonstrated enhanced transferability and anti-purification capabilities by employing diffusion-based adversarial purification models for adaptive attacks. Although DiffPGD offers an accelerated version achieved through gradient approximation, a trade-off between performance and computational cost remains necessary due to the inherent expense of diffusion models. Therefore, developing a less computationally intensive method is essential for practical applications.

Our method aims to address the significant computational overhead of typical approaches while maintaining both transferability and anti-purification capabilities for adversarial perturbations. Based on the hypothesis presented at the end of Section~\ref{dbp}, we suggest that the noise addition process is what makes diffusion-based purification robust against adversarial samples. Previous methods such as DiffPGD achieves transferability and anti-purification by incorporating the entire diffusion model into the PGD framework. However, we observe that the time-dependent classifier, $f_\phi(\bm{x}_t,t)$, inherently utilizes the same noising process, which could provide the necessary knowledge we hypothesized. To effectively integrate this insight into the PGD algorithm in a computationally efficient manner, we propose a novel loss function that can simultaneously attack both classification and purification.

Since time-dependent classifiers can guide the reverse diffusion process, we naturally consider an anti-guidance approach. By redirecting the guidance toward an incorrect class (as shown in Figure~\ref{fig:score target} ), we can attack both the purification mechanism that utilizes the diffusion model and the classification task itself.

Fundamentally, our optimization goal is to minimize the log likelihood of the true label $y$ given the noised image $\bm{x}_t$ by perturbing the clean image $\bm{x}$ with a perturbation budget $\gamma$. This can be formulated as:

\begin{equation}
    \bm{\delta}=\argmin_{\bm{\delta}} \log p_\phi(y\vert\bm{x}_{adv}^t) \text{ , s.t., }
     d(\bm{x}_{adv} , \bm{x}) \leq \gamma,
\end{equation}
where $\bm{x}_t$ is derived from $\bm{x}$ according to Equation~\ref{forward process}.

Since we do not have access to the ground truth distribution $p_\phi(y\vert \bm{x}_{adv}^t)$, we instead attack its approximation using the time-dependent classifier mentioned earlier. The corresponding optimization problem becomes:
\begin{equation}
    \bm{\delta}= \argmin_{\bm{\delta}} \log f_\phi(\bm{x}_{adv}^t,t) \text{, s.t., } d(\bm{x}_{adv} , \bm{x}) \leq \gamma,
    \label{scoreloss}
\end{equation} 
where we denote $\log f_\phi(\bm{x}_{adv}^t,t)$ as $\mathcal{L}_s$ for concise notation. We then change the sign of $\mathcal{L}_s$, and employ PGD to solve this optimization problem by iteratively optimizing $\bm{\delta}$:

\begin{equation}
\bm{\delta}^{(i+1)}=\mathcal{P}_{\gamma}\left(\bm{\delta}^{(i)}+\eta \cdot \text{sign}\nabla_{\bm{\delta}^{(i)}}\mathcal{L}_s\right),
    \label{pgdforls}
\end{equation}
where $\mathcal{P}$ is the projection operation onto an $\ell_p$ norm ball $B_p$ with radius $\gamma$. This projection ensures that $\bm{x}_{adv}$ satisfies the constraint $d(\bm{x}_{adv} , \bm{x}) \leq \gamma$. Common choices for $p$ are $p=2$ and $p=\infty$, corresponding to the $\ell_2$ and $\ell_\infty$ distance metrics, respectively. Here, $i$ represents the current iteration step and $\eta$ is the step size for each iteration. The full algorithm of ScorePGD can be found in Algorithm~\ref{algo:score}.

By optimizing $\mathcal{L}_s$ by our ScorePGD, the guidance direction to the image is redirected, which can protect from unauthorized image editing. Visualization is shown in the bottom part of Figure~\ref{fig:teaser}.

\begin{algorithm}[t]
\caption{Universal ScorePGD (U-ScorePGD)}
\label{algo:full}
\begin{algorithmic}
\Require Target threat model $f_\theta$, original image $\bm{x}$, time-dependent classifier $f_\phi(\bm{x}_t,t)$, predefined timestep $t$ for the time-dependent classifier, iteration $n$, stepsize $\eta$ and bound value $\gamma$,

\State $\bm{x}_{adv}=\bm{x}$
\For{$t=0,1,2,\dots,n-1$}
\State $\mathcal{L}_c$ = $\mathcal{L}_c(f_\theta(\bm{x}_{adv},y))$ \Comment{Calculate corresponding loss for the threat model}
\State $\bm{\epsilon}\sim \mathcal{N}(\mathbf{0},\mathbf{I})$
\State $\bm{x}_{adv}^t=\sqrt{\alpha_t}\bm{x}_{adv}+\sigma_t\bm{\epsilon}$ \Comment{Sample $\bm{x}_{adv}^t$ for each PGD iteration by Equation~\ref{forward process}}
\State $\mathcal{L}_s = \log f_\phi(\bm{x}_{adv}^t,t) $ \Comment{Calculate noise score loss by Equation~\ref{scoreloss}}
\State $\mathcal{L}_t = \mathcal{L}_c - \mathcal{L}_s$ \Comment{Combine two loss by Equation~\ref{combineloss}}
\State $\bm{\delta}^{(i+1)}=\mathcal{P}_{\gamma}\left(\bm{\delta}^{(i)}+\eta \cdot \text{sign}\nabla_{\bm{\delta}^{(i)}}\mathcal{L}_t\right)$   \Comment{Update the value of $\delta$ by Equation~\ref{pgduqdate}}
\State $\bm{x}_{adv} = \bm{x}_{adv}+\bm{\delta}$ 
\EndFor
\State\Return $x_{adv}$
\end{algorithmic}
\end{algorithm}
\subsection{Universal Score PGD}

For any computer vision tasks, suppose we have an image sample $\bm{x} \in \mathcal{X} \coloneqq \mathbb{R}^{H \times W \times C}$, where $H$, $W$, and $C$ represent the height, width, and number of channels of the image, respectively, and ground truth $y \in \mathcal{Y}$ denote its ground truth label, where $\mathcal{Y}$ is the set of all possible labels.

For any model $f: \mathcal{X} \to \mathcal{Y}$, given $\bm{x}$ and $y$, our objective is to craft an adversarial example $\bm{x}_{adv}$ such that:
\begin{equation}
    f(\bm{x}_{adv})\neq y \text{ and } f(D(\bm{x}_{adv}))\neq y \text{, s.t., }d(\bm{x}_{adv},\bm{x})\leq \gamma,
\end{equation}

where $\bm{x}_{adv} = \bm{x} + \bm{\delta}$, $D$ represents a diffusion model used for purification, $d$ is a distance metric, and $\gamma$ is the perturbation budget. 

To allow us manipulating our method in white-box setting, we incorporate a surrogate model corresponding to the task that operates on non-noisy images. We can then formulate an optimization problem:
\begin{equation}
    \bm{\delta}=\argmax_{\bm{\delta}} \mathcal{L}_c(f_\theta (\bm{x}_{adv}),y) \text{, s.t., } d(\bm{x}_{adv} , \bm{x}) \leq \gamma,
    \label{crossentropyloss}
\end{equation}
where $\mathcal{L}_c$ is the corresponding task's training loss.

To avoid updating the perturbation $\bm{\delta}$ twice in each PGD iteration, we combine the two optimization problems from Equation~\ref{scoreloss} and Equation~\ref{crossentropyloss} into a single objective:
\begin{equation}
    \bm{\delta}=\argmax_{\bm{\delta}} \mathcal{L}_c(f_\theta (\bm{x}_{adv}),y) - \mathcal{L}_s\text{, s.t., } d(\bm{x}_{adv} , \bm{x}) \leq \gamma.
    \label{combineloss}
\end{equation}

We denote $\mathcal{L}_c(f_\theta (\bm{x}_{adv}),y) - \mathcal{L}_s$ as $\mathcal{L}_t$ and solve the optimization problem in Equation~\ref{combineloss} using PGD, similar to Equation~\ref{pgdforls}:
\begin{equation}
\bm{\delta}^{(i+1)}=\mathcal{P}_{\gamma}\left(\bm{\delta}^{(i)}+\eta \cdot \text{sign}\nabla_{\bm{\delta}^{(i)}}\mathcal{L}_t\right).
    \label{pgduqdate}
\end{equation}

The full algorithm of our proposed method U-ScorePGD (as shown in Figure~\ref{fig:pipeline}) can be found in Algorithm~\ref{algo:full}.

\section{Experiments and Results}

\subsection{Experiment Settings}

We compared our proposed method with PGD~\cite{madry2018towards}, and DiffPGD~\cite{xue2023diffusion}. We follow the convention of adversarial attack where using a subset~\cite{lin2019nesterov} of ImageNet~\cite{deng2009imagenet} to promptly test the performance of different attack methods. The main target classifier to be attacked across this paper is ResNet50~\cite{he2016deep}, and we also use ResNet101, ResNet18, and WideResNet (WRN)~\cite{zagoruyko2016wide} for the transferability test. 

The time-dependent classifier~\cite{dhariwal2021diffusion} for the implementation of our method was pretrained on ImageNet with resolution $256\times256$.

We present the quantitative visual evaluation of adversarial samples with a series of Image Quality Assessment (IQA): Peak Signal-to-Noise Ratio (PSNR), Structure Similarity Index Metric (SSIM)~\cite{wang2004image}, and Learned Perceptual Image Patch Similarity (LPIPS)~\cite{zhang2018unreasonable}.
We utilize these IQA metrics to qualitatively measure the differences between original and purified images. Notably, worth IQA scores indicate stronger anti-purification capabilities, as they reflect greater preservation of adversarial features during the purification process.

For a fair comparison, all PGD-form attacks are configured with $\mathcal{L}_\infty$ bounds $\gamma\in \{8/255, 16/255\}$. For the DiffPGD, we fully follow the default setting reported in its paper. In our evaluation metrics, we will report attack success rate (ASR) for evaluating attack performance. All the experiments below are run on a single NVIDIA F40 GPU.

\begin{table}[t!]
\small
\centering
\caption{Adversarial samples generated by PGD, DiffPGD and ours. The attacks are operated on ResNet50. The white-box scenario for surrogate models are represented with \colorbox{gray!20}{gray backgrounds}. The best result is in \textbf{bold}.}
\label{tab:imagenet}
\setlength{\tabcolsep}{0.5mm}
\resizebox{\linewidth}{!}{%
    \begin{tabular}{ccccccc|ccccccc}
    \toprule
     $\eta$&Method & ResNet50  & ResNet101 & ResNet18 & WRN50 & WRN101&$\eta$&Method & ResNet50  & ResNet101 & ResNet18 & WRN50 & WRN101\\
     \midrule
         \multirow{ 4}{*}{$\dfrac{16}{255}$}&PGD & \cellcolor{gray!20}\textbf{100}\% & 84.3\% & 71.4\% & 79.0\% & 69.2\% & \multirow{ 4}{*}{$\dfrac{8}{255}$}&PGD & \cellcolor{gray!20}\textbf{100}\% & {62.7}\% & {48.7}\% & {55.4}\% & {48.6}\%\\
 
         &DiffPGD~\cite{xue2023diffusion}, 
         & \cellcolor{gray!20}\textbf{100}\% & 84.1\% & 77.5\% & 79.5\% & 74.8\% & &DiffPGD~\cite{xue2023diffusion}, & \cellcolor{gray!20}99.0\%& 61.9\% & 54.3\% & 54.1\% & 49.5\% \\
         \cmidrule{2-7} \cmidrule{9-14}
         &ScorePGD (Ours) &66.4\% &65.6\% &80.2\% &63.5\% & 62.1\%    & &ScorePGD (Ours) &38.2\% &34.3\% &51.8\% &31.6\% & 30.7\%    \\
         &U-ScorePGD (Ours) &\cellcolor{gray!20}99.6\% & \textbf{89.9}\% & \textbf{85.6}\% & \textbf{87.6}\% & \textbf{81.8}\% &  &U-ScorePGD (Ours) & \cellcolor{gray!20}{99.6}\% & \textbf{69.1}\% & \textbf{62.7}\% & \textbf{62.3}\% & \textbf{59.0}\% \\
        \bottomrule
    \end{tabular}}
\end{table}

\subsection{Attack Performance on Unprotected Classifier}
Table \ref{tab:imagenet} demonstrates the superior performance of our method on unprotected classifiers under different perturbation bounds. In white-box settings for the surrogate model (highlighted with gray background), our U-ScorePGD method maintains comparable performance with other baseline approaches. Notably, the pure Score-based PGD shows some performance degradation in white-box scenarios, which is reasonable since the method lacks knowledge of the classifier's inductive bias, making this setting effectively black-box for the threat task.

In black-box transfer scenarios, our U-ScorePGD method shows significant advantages compared to other baselines by effectively combining with the classification task's inductive bias. This leads to consistently optimal performance across all tested architectures. Specifically, at $\gamma=16/255$, U-ScorePGD achieves 89.9\%, 85.6\%, 87.6\%, and 81.8\% ASR on ResNet101, ResNet18, WRN50, and WRN101 respectively, outperforming both standard PGD and DiffPGD by substantial margins. Similar improvements are observed at the more restrictive $\gamma=8/255$ bound, where our method maintains its advantage.

\begin{table}[t!]
\caption{Adversarial samples generated by PGD, DiffPGD and ours against adversarial purification: (+P) means the classifier is protected by an adversarial purification module. The attacks are operated on ResNet50. The white-box scenario for surrogate models are represented with \colorbox{gray!20}{gray backgrounds}. The best result is in \textbf{bold}, the second best is in \underline{underline} and the full knowledge setting regarding purification is denoted by \textcolor{lightgray}{light gray}.}
    \centering
    \setlength{\tabcolsep}{0.5mm}
    \resizebox{\textwidth}{!}{
    \begin{tabular}{ccccccc|ccccccc}
    \toprule
    $\eta$& (+P)Method  & ResNet50  &ResNet101 & ResNet18 & WRN50 & WRN101 & $\eta$& Method & ResNet50  &ResNet101 & ResNet18 & WRN50 & WRN101\\
     \midrule
         \multirow{ 4}{*}{$\dfrac{16}{255}$}&PGD & \cellcolor{gray!20}35.1\% & 14.3\% & 23.0\% & 12.3\% & 12.5\% &\multirow{ 4}{*}{$\dfrac{8}{255}$}&PGD  & \cellcolor{gray!20}19.6\% & 10.5\% & 20.1\% & 9.3\% & 9.5\%\\
         &DiffPGD~\cite{xue2023diffusion} 
         & \cellcolor{gray!20}\textcolor{lightgray}{93.4\%} & {33.7}\% & {38.7}\% & {33.7}\% & {30.7}\% & &DiffPGD~\cite{xue2023diffusion} & \cellcolor{gray!20}\textcolor{lightgray}{69.4\%} & {21.5}\% & {28.1}\% & {18.7}\% & {17.2}\% \\
         \cmidrule{2-7} \cmidrule{9-14}
         & ScorePGD (Ours) &\textbf{66.4}\% & \textbf{64.6}\% &\textbf{75.4}\% &\textbf{62.1}\% & \textbf{61.7}\% && ScorePGD (Ours) &\textbf{39.0}\% & \textbf{37.8}\% &\textbf{50.9}\% &\textbf{33.1}\% & \textbf{34.8}\% \\
         &U-ScorePGD (Ours) & \cellcolor{gray!20}\underline{56.4}\% & \underline{36.5}\% & \underline{49.4}\% & \underline{36.9}\% & \underline{34.3}\% &  &U-ScorePGD (Ours) & \cellcolor{gray!20}\underline{32.7}\% & \underline{21.6}\% & \underline{33.8}\% & \underline{20.1}\% & \underline{20.7}\% \\
        \bottomrule
    \end{tabular}}
    \label{tab:imagenet robust}
\end{table}

\subsection{Attack Performance on Protected Classifier}

Table \ref{tab:imagenet robust} demonstrates our method's superior performance against classifiers protected by adversarial purification. In white-box settings, our pure Score-based method consistently achieves the best results, significantly outperforming both PGD and DiffPGD approaches. At $\gamma=16/255$, ScorePGD attains 66.4\%, 64.6\%, 75.4\%, 62.1\%, and 61.7\% ASR against protected ResNet50, ResNet101, ResNet18, WRN50, and WRN101 respectively.

Notably, while U-ScorePGD was optimal for unprotected classifiers, the pure ScorePGD method proves superior against protected models. This suggests that score-based approaches are inherently more effective at bypassing adversarial purification mechanisms, likely by leveraging natural image manifold properties that purification defenses attempt to exploit.

Even at the more constrained $\gamma=8/255$ bound, our ScorePGD method maintains its advantage, demonstrating robustness across different perturbation constraints when targeting protected models. It is worth mentioning that DiffPGD has knowledge of both ResNet50 and the purification model, giving it a significant advantage compared to PGD and our method, and thus achieving better results. The fully white-box setting is marked in light gray in the Table~\ref{tab:imagenet robust}.

Table \ref{tab:IQA origin vs purified image} presents image quality metrics that demonstrate our Score-based method's effectiveness at disrupting purification defense mechanisms. The key insight from these results is that while different methods may produce seemingly similar adversarial examples initially, our method specifically compromises the purification process itself.

\begin{table}[t!]
    \centering
    \caption{Comparison of our proposed method with baseline methods in terms of Image Quality
Assessment (IQA) metrics. The results at bottom are comparing between clean image and purified adversarial image and results inside the bracket are comparing between purified clean image and purified adversarial image. Arrows ($\uparrow$/$\downarrow$) indicate the direction of better attack performance for each image quality metric.}
\resizebox{\linewidth}{!}{
    \begin{tabular}{c|ccc|ccc}
    \toprule
    & &  $\gamma=16/255$& &&$\gamma=8/255$  &  \\
    \midrule
    Method& SSIM $\downarrow$ &  PSNR $\downarrow$& LPIPS $\uparrow$& SSIM $\downarrow$ &  PSNR $\downarrow$& LPIPS $\uparrow$ \\
    \midrule
        PGD & $0.8661^{(0.8745)}$& $29.57^{(30.03)}$&$0.1521^{(0.1472)}$ &$0.8748^{(0.8869)}$&  $29.74^{(30.28)}$&$0.1510^{(0.1301)}$\\
   
        DiffPGD~\cite{xue2023diffusion} &$0.8595^{(0.8661)}$&$29.40^{(29.81)}$&$0.1622^{(0.1634)}$ & $0.8732^{(0.8846)}$ & $29.70^{(30.21)}$& $0.1531^{(0.1340)}$
        \\
        \midrule
        ScorePGD (Ours) & $\textbf{0.7822}^{(\textbf{0.7837})}$& $\textbf{26.58}^{(\textbf{26.81})}$ &$\textbf{0.3029}^{(\mathbf{0.2978})}$& $\textbf{0.8414}^{(\textbf{0.8491})}$& 
        $\textbf{28.48}^{(\textbf{28.86})}$& $\textbf{0.2139}^{(\textbf{0.2027})}$
        \\
        U-ScorePGD (Ours) & $\underline{0.8522}^{(\underline{0.8592})}$& $\underline{28.80}^{(\underline{29.19})}$ &$\underline{0.1844}^{(\underline{0.1799})}$& $\underline{0.8689}^{(\underline{0.8803})}$& $\underline{29.38}^{(\underline{29.89})}$& $\underline{0.1640}^{(\underline{0.1444})}$
        \\
        \bottomrule
    \end{tabular}
    }
    \label{tab:IQA origin vs purified image}
\end{table}

For $\gamma=16/255$, our ScorePGD method achieves the lowest SSIM (0.7822), lowest PSNR (26.58), and highest LPIPS (0.3029) when comparing original clean images with purified adversarial images. These metrics, substantially worse than other methods, indicate that purification largely fails when applied to our adversarial examples. The similar values in parentheses further confirm that even when comparing purified clean images to purified adversarial images, significant differences remain.

U-ScorePGD also demonstrates strong capabilities in disrupting purification mechanisms, ranking second across all metrics. This pattern holds at the lower perturbation bound $\gamma=8/255$, where Score continues to show the strongest ability to compromise purification defenses.

These results explain why our Score method achieves higher attack success rates against protected classifiers --- it fundamentally undermines the purification process that these defenses rely on.

\subsection{Runtime Comparison}
\begin{wraptable}{r}{0.6\textwidth}
    \centering
    \small
    \setlength{\tabcolsep}{0.5mm}
    \caption{Runtime comparison between ours and DiffPGD.}
\resizebox{\linewidth}{!}{\begin{tabular}{c|ccc|ccc}
\toprule
 & &  $256\times256$& &&$512\times512$  & \\ \midrule
Method          & $n=10$   & $n=20$& $n=50$&  $n=10$   & $n=20$& $n=50$\\ \midrule

DiffPGD~\cite{xue2023diffusion}   &   $\sim 5$  s             & $\sim 10$  s  &  $\sim$ 23 s &$\sim 10$  s             & $\sim 20$  s  &  $\sim$ 50 s \\ \midrule
ScorePGD (Ours)       & $\sim\textbf{0.4}$ s   & $\sim\textbf{0.8}$ s& $\sim$\textbf{2} s& $\sim\textbf{0.6}$ s   & $\sim\textbf{1.2}$ s& $\sim$\textbf{ 3.1} s\\
U-ScorePGD (Ours)       & $\sim\underline{0.6}$ s   & $\sim\underline{1.1}$ s& $\sim$\underline{2.7} s& $\sim\underline{0.8}$ s   & $\sim\underline{1.5}$ s& $\sim$\underline{ 3.8} s

\\ \bottomrule

\end{tabular}
}    
    \label{tab:time}
\end{wraptable}

Table \ref{tab:time} demonstrates the significant computational efficiency of our proposed methods compared to DiffPGD. Our ScorePGD approach achieves remarkable speedups across all tested configurations. For $256\times256$ images with $n=50$ iterations, ScorePGD completes in just around 2 seconds compared to DiffPGD's around 23 seconds, representing more than a $10\times$ improvement. Similar efficiency gains are observed for larger $512\times512$ images, where ScorePGD requires only around 3.1 seconds versus DiffPGD's around 50 seconds. Even our U-ScorePGD variant, while slightly slower than ScorePGD, maintains substantial speed advantages over DiffPGD. These results demonstrate that beyond superior attack performance, our methods also offer practical computational benefits that enable more efficient adversarial testing.

\section{Conclusion}
In this paper, we proposed a novel approach to enhance the transferability of adversarial attacks while drastically reducing computational costs. By leveraging the score of time-dependent classifiers from classifier-guided diffusion models, our method effectively incorporates noised data distribution knowledge without requiring the complete diffusion machinery.
Our experiments confirmed that our ScorePGD and U-ScorePGD methods achieve superior performance across diverse scenarios. Against unprotected classifiers, U-ScorePGD consistently outperforms both standard PGD and DiffPGD in black-box transfer settings. For protected classifiers, our pure ScorePGD method demonstrates remarkable effectiveness at undermining adversarial purification defenses.
Most significantly, our approach achieves these superior results with an order of magnitude reduction in computational requirements—completing in just 2 seconds compared to DiffPGD's 23 seconds for 256×256 images. These findings validate our hypothesis that efficient incorporation of noised data distribution knowledge, rather than the complete diffusion process, is the critical factor for developing highly transferable adversarial attacks. Our work establishes a new efficiency-performance balance for transferable attacks and enables more practical robustness evaluation in resource-constrained environments. 

\newpage
\bibliographystyle{unsrt}
\bibliography{main}
\newpage

\appendix
\section{Appendix}

\subsection{Notation}
In this section, we provide the detailed description of notations we used in the main paper:
\begin{table}[h]
    \centering
    \resizebox{0.7\linewidth}{!}{
    \begin{tabular}{cl}
    \toprule
    \textbf{Notation} & \textbf{Description} \\
    \midrule
      $f_\phi(x_t,t)$ & time-dependent classifier trained on noised image \\
  $q(x_t|x_0)$& forward diffusion process in diffusion model\\
$p_\theta(x_{t-1}|x_t)$& backward diffusion process parameterized by $\theta$\\
$\alpha,\sigma$&  diffusion coefficients in a diffusion model\\
$\epsilon_\theta$& U-Net of diffusion model used to predict added noise\\
       $n$ & number of iterations of PGD attack \\
      $\gamma$ & maximum $l_\infty$ norm of adversarial attacks \\
       $\eta$ & step size of adversarial attacks \\
       $x^t_{adv}$ & diffused adversarial image\\
       $\delta$ & adversarial perturbation \\
       $B_p$ & $l_p$ ball \\ 
       $D$ & diffusion model for purification \\
       \bottomrule
    \end{tabular}}
\end{table}

\subsection{Implementation Details}
We implement our method with Pytorch~\cite{paszke2019pytorch}.
The random seed of PyTorch’s generator is fixed as 3407~\cite{picard2021torch} for reproducibility.
We leverage diffusion model and time-dependent classifier in Guided-diffusion~\cite{dhariwal2021diffusion} from their GitHub repository\footnote{\url{https://github.com/openai/guided-diffusion}}. We also conduct a series of IQAs to evaluate the quality of purified images. To be specific, we leverage:
\begin{itemize}
\item PSNR: Range set to 1, aligning with PyTorch's image transformation.
\item {SSIM}\footnote{\url{https://github.com/Po-Hsun-Su/pytorch-ssim}}~\cite{wang2004image}: Gaussian kernel size set to 11.
\item {LPIPS}\footnote{\url{https://github.com/richzhang/PerceptualSimilarity}}~\cite{zhang2018unreasonable}: Utilizing VGG \cite{simonyan2014very} as the surrogate model.
\end{itemize}

Regarding the main results we presented in the main paper, for our method, we implement it in $\ell_\infty$-norm PGD, with the number of iterations to 10, $\gamma$ to \{16/255, 8/255\}, and step size to \{2/255, 1/255\}. For DiffPGD, we use their version 2, which is the most frequently used variant. For purification method, we fully follow the setting reported in DiffPGD's paper. All the experiments are run on a subset~\cite{lin2019nesterov} of ImageNet~\cite{deng2009imagenet}, which is a convention in adversarial attack.

It is worth noting that we borrow the implementation of PSNR from {TorchEval}\footnote{\url{https://pytorch.org/torcheval/stable/}}.
Unless mentioned, all reproducibility-related things follow the above.

\subsection{Ablation on Diffusion Timestep $t$ in Time-dependent Classifier in U-ScorePGD}

We are testing different $t $ in the time-dependent classifier for U-ScorePGD in Table~\ref{ablation uscore} and Table~\ref{ablation uscore with p}, where Table~\ref{ablation uscore} is testing on classifier without protection from purification method and Table~\ref{ablation uscore with p} is testing on classifier with protection from purification method. The surrogate model is Resnet50 for both Table~\ref{ablation uscore} and Table~\ref{ablation uscore with p}.

\begin{table}[ht]
    \centering
    \caption{Attack success rate of U-ScorePGD in different $t$ in time-dependent classifier. The white-box scenario for surrogate models are represented with \colorbox{gray!20}{gray backgrounds}. The best result is in \textbf{bold}.}
    \begin{tabular}{c|ccccc}
    \toprule
   Victim model& ResNet50&ResNet101&ResNet18&WRN50&WRN101\\ 
   \midrule
     $t=20$   & \cellcolor{gray!20}{99.6}\% & \textbf{89.9}\% & \textbf{85.6}\% & \textbf{87.6}\% & \textbf{81.8}\% \\
    $t=40$&   \cellcolor{gray!20}{\textbf{99.7}}\%&  88.7\%& 85.2\%& 86.7\%& 80.0\%\\
    $t=60$&  \cellcolor{gray!20}{\textbf{99.7}\%}&88.8\%& 85.4\%&85.5\%&79.3\%\\
   \bottomrule
    \end{tabular}
    
    \label{ablation uscore}
\end{table}

In Table~\ref{ablation uscore}, $t=20$ preform the best in general, with a slightly lower white-box attack success rate 99.6\% and higher black-box attack success rate 89.9\%, 85.6\%, 87.6\% and 81.8\% in ResNet101, ResNet18, WideResNet50 and WideResNet101 respectively.

\begin{table}[ht]
    \centering
    \caption{Attack success rate of U-ScorePGD against adversarial purification in different $t$ in time-dependent classifier. 
(+P) means the classifier is protected by an adversarial purification module. The white-box scenario for surrogate models are represented with \colorbox{gray!20}{gray backgrounds}. The best result is in \textbf{bold}.}
    \begin{tabular}{c|ccccc}
    \toprule
   (+P)Victim model& Resnet50&Resnet101&Resnet18&WRN50&WRN101\\ 
   \midrule
     $t=20$   & \cellcolor{gray!20}{56.4}\% & {36.5}\% & \textbf{49.4}\% & \textbf{36.9}\% & \textbf{34.3}\% \\
    $t=40$&   \cellcolor{gray!20}{\textbf{57.8}\%}&  \textbf{37.5}\%& 49.3\%& 35.6\%& 32.8\%\\
    $t=60$&  \cellcolor{gray!20}{55.9\%}&35.8\%&\textbf{49.4}\%&33.9\%&32.7\%\\
   \bottomrule
    \end{tabular}
    
    \label{ablation uscore with p}
\end{table}

In Table~\ref{ablation uscore with p}, $t=20$ and $t=40$ preform similarly, where $t=40$ have a better white-box attack success rate 57.8\% and higher black-box attack success rate 37.5\% in ResNet101, $t=20$ have better attack success rate 49.4\%, 36.9\% and 34.3\% in ResNet18, WideResNet50 and WideResNet101 respectively. By these results, we decide to use $t=20$ for U-ScorePGD. Worth to note that, all the result in Table~\ref{ablation uscore} and Table~\ref{ablation uscore with p} are better than DiffPGD, except white-box setting, which can be found in the main paper.

\subsection{Ablation on Diffusion Timestep $t$ in Time-dependent Classifier in ScorePGD}

We are testing different diffusion timestep $t$ in the time-dependent classifier for ScorePGD in Table~\ref{ablation score} and Table~\ref{ablation score with p}, where Table~\ref{ablation score} is testing on classifier without protection from purification method and Table~\ref{ablation score with p} is testing on classifier with protection from purification method.

\begin{table}[ht]
    \centering
    \caption{Attack success rate of ScorePGD in different $t$ in time-dependent classifier. The best result is in \textbf{bold}.}
    \begin{tabular}{c|ccccc}
    \toprule
    Victim model& Resnet50&Resnet101&Resnet18&WRN50&WRN101\\ 
   \midrule
     $t=20$   & \textbf{66.4}\% &\textbf{65.6}\% &\textbf{80.2}\% &\textbf{63.5}\% & \textbf{62.1}\% \\
    $t=40$&   64.6\%&  62.9\%& 78.6\%& 63.3\%& 60.1\%\\
    $t=60$&  62.0\%&60.9\%&76.1\%&59.1\%&55.3\%\\
   \bottomrule
    \end{tabular}
    
    \label{ablation score}
\end{table}

\begin{table}[ht]
    \centering
    \caption{Attack success rate of ScorePGD against adversarial purification in different $t$ in time-dependent classifier. (+P) means the classifier is protected by an adversarial purification module.
    The best result is in \textbf{bold}.}
    \begin{tabular}{c|ccccc}
    \toprule
   (+P)Victim model& Resnet50&Resnet101&Resnet18&WRN50&WRN101\\ 
   \midrule
     $t=20$   & {66.4}\% & \textbf{64.6}\% &\textbf{75.4}\% &\textbf{62.1}\% & \textbf{61.7}\%\\
    $t=40$&   \textbf{66.7}\%&  {63.4}\%& 74.3\%& 62.0\%& 60.8\%\\
    $t=60$&  63.9\%&59.8\%&72.0\%&59.0\%&57.0\%\\
   \bottomrule
    \end{tabular}
    
    \label{ablation score with p}
\end{table}

In Table~\ref{ablation score}, $t=20$ preform the best in general, with black-box attack success rate 66.4\%, 65.6\%, 80.2\%, 63.5\% and 62.1\% in ResNet50, ResNet101, ResNet18, WideResNet50 and WideResNet101 respectively.
In Table~\ref{ablation score with p}, also $t=20$ preform the best in general, with black-box attack success rate 66.4\%, 64.6\%, 75.4\%, 62.1\% and 61.7\% in ResNet50, ResNet101, ResNet18, WideResNet50 and WideResNet101 respectively. $t=40 $ preform slightly better than $t=20$ in ResNet50 with attack success rate 66.7\%. By the above results, we decide to choose $t=20$ for ScorePGD.

This is an interesting phenomenon that relatively smaller $t$ has better performance.

This phenomenon occurs because images become increasingly noisy as $t$ increases. At larger timesteps, the Gaussian noise added during the forward diffusion process overwhelms the adversarial perturbation, resulting in $\sigma_t\epsilon \gg \sqrt{\alpha_t}\delta$. This fundamental imbalance prevents effective optimization of the perturbation $\delta$. We validate this claim through empirical observation: when $t$ is large, the score loss $\mathcal{L}_s$ fails to change during PGD optimization iterations. The same limitation applies to U-ScorePGD.

\subsection{Results on $\ell_2$-norm Attacks}
\label{l2 setting}

Since the $\ell_2$ distance between an adversarial image and a natural image is defined as
$\vert\vert x_{adv}-x\vert\vert_2=\sqrt{\sum_{i=1}^N(x_{adv}-x)^2}$, where $N$ is the dimension of input $x$ and $x_{adv}$. We need a pertubration budget $\gamma\sqrt{N}$ in $\ell_2$ to perform just like $\vert\vert x_{adv}-x\vert\vert_\infty\leq\gamma$ in $\ell_\infty$. Hence, we test PGD, DiffPGD and ours in $\ell_2$, with $\gamma=16/255*244$, $\eta=2/255*244$ and $n=10$.

The results of $\ell_2$ are shown in Table~\ref{l2} and Table~\ref{l2 with p}, where Table~\ref{l2} is testing
on classifier without protection from purification method and Table~\ref{l2 with p} is testing on classifier with
protection from purification method. The surrogate model is ResNet50 for both Table~\ref{l2} and Table~\ref{l2 with p}.

In Table~\ref{l2}, our method U-ScorePGD perform the best in all architectures and in both white-box and black-box. Showing our method can easily extend to $\ell_2$-based adversarial attack. In Table~\ref{l2 with p}, ScorePGD has the best performance and U-ScorePGD has the second best performance, which align when the results in $\ell_\infty$-based adversarial attack in the main paper. These results are extremely supportive result for our method has more transferability and anti-purify ability.

\begin{table}[ht]
    \centering
    \caption{Adversarial samples generated by PGD, DiffPGD and ours. The attacks are operated on
ResNet50. The white-box scenario for surrogate models are represented with \colorbox{gray!20}{gray backgrounds}. The best result is in \textbf{bold}.}
    \begin{tabular}{c|ccccc}
    \toprule
   Method & Resnet50&Resnet101&Resnet18&WRN50&WRN101\\ 
   \midrule
     PGD   & \cellcolor{gray!20}{\textbf{100}\%} & 82.7\% &71.2\% &80.6\% & 71.0\%\\
    DiffPGD&   \cellcolor{gray!20}{99.7\%}&82.2\%&74.8\%&78.3\%&73.5\%\\ 
    \midrule
    U-ScorePGD&  \cellcolor{gray!20}{\textbf{100}\%}&\textbf{88.6}\%&\textbf{83.8}\%&\textbf{86.7}\%&\textbf{82.0}\%\\
   \bottomrule
    \end{tabular}
    
    \label{l2}
\end{table}

\begin{table}[ht]
    \centering
    \caption{Adversarial samples generated by PGD, DiffPGD and ours against adversarial purification:
(+P) means the classifier is protected by an adversarial purification module. The attacks are operated on
ResNet50. The white-box scenario for surrogate models are represented with \colorbox{gray!20}{gray backgrounds}. The best result is in \textbf{bold}, the second best is in \underline{underline} and the full knowledge setting regarding purification is denoted by \textcolor{lightgray}{light gray}.}
    \begin{tabular}{c|ccccc}
    \toprule
   (+P)Method & Resnet50&Resnet101&Resnet18&WRN50&WRN101\\ 
   \midrule
     PGD   & \cellcolor{gray!20}{29.1\%}&  14.7\%& 20.9\%& 13.7\% & 11.8\%\\
    DiffPGD&   \cellcolor{gray!20}{\textcolor{lightgray}{86.6\%}}&  34.6\%& 39.9\%& 33.7\%& 29.7\%\\ 
    \midrule
    ScorePGD&  \textbf{70.0}\%&\textbf{69.6}\%&\textbf{78.4}\%&\textbf{68.6}\%&\textbf{67.0}\%\\
    U-ScorePGD&  \cellcolor{gray!20}{\underline{52.6}\%}&\underline{37.6}\%&\underline{48.8}\%&\underline{36.8}\%&\underline{36.0}\%\\
   \bottomrule
    \end{tabular}
    
    \label{l2 with p}
\end{table}

\subsection{Results on Vision Transformers}
Apart from Convolutional-based network, we would like to try transformer-based networks as well. We test on ViT-b and Swin-b, the result are shown in Table~\ref{vit} and Table~\ref{vit with p}, where Table~\ref{vit} is testing on classifier without protection from purification method and Table~\ref{vit with p} is testing on classifier with protection from purification method. The surrogate model is ViT-b for both Table~\ref{vit} and Table~\ref{vit with p}.

\begin{figure*}
\centering
\resizebox{\linewidth}{!}{
\begin{tabular}[b]{ccccc}
\includegraphics[width=0.19\textwidth]{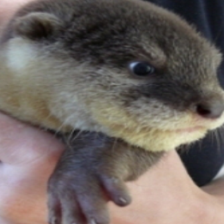}
     &
     \hspace{-1em}
    \includegraphics[width=0.19\textwidth]{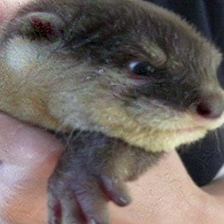}
     &
    \hspace{-1em}
    \includegraphics[width=0.19\textwidth]{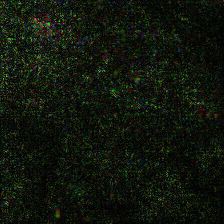}
     & 
     
    \hspace{-1em}
        \includegraphics[width=0.19\textwidth]{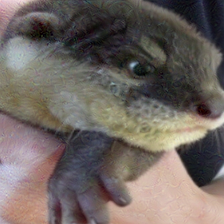}
        &
        \hspace{-1em}
        \includegraphics[width=0.19\textwidth]{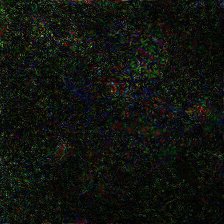}
     \\
     \includegraphics[width=0.19\textwidth]{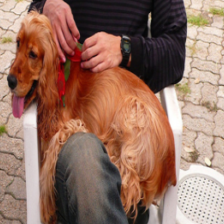}
     &
     \hspace{-1em}
    \includegraphics[width=0.19\textwidth]{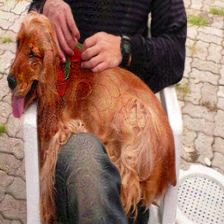}
     &
    \hspace{-1em}
    \includegraphics[width=0.19\textwidth]{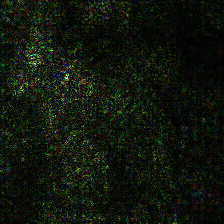}
     & 
     
    \hspace{-1em}
        \includegraphics[width=0.19\textwidth]{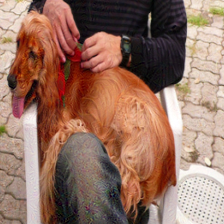}
        &
        \hspace{-1em}
        \includegraphics[width=0.19\textwidth]{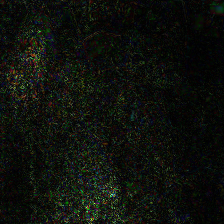}
     \\\includegraphics[width=0.19\textwidth]{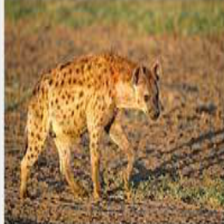}
     &
     \hspace{-1em}
    \includegraphics[width=0.19\textwidth]{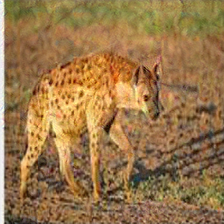}
     &
    \hspace{-1em}
    \includegraphics[width=0.19\textwidth]{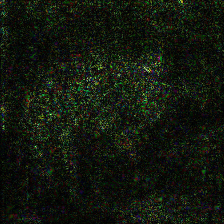}
     & 
     
    \hspace{-1em}
        \includegraphics[width=0.19\textwidth]{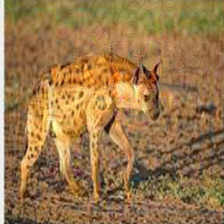}
        &
        \hspace{-1em}
        \includegraphics[width=0.19\textwidth]{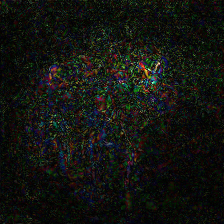}
     \\\includegraphics[width=0.19\textwidth]{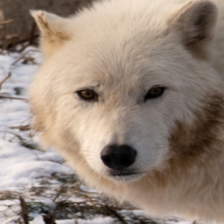}
     &
     \hspace{-1em}
    \includegraphics[width=0.19\textwidth]{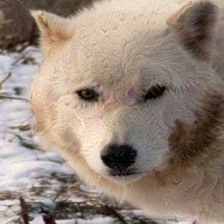}
     &
    \hspace{-1em}
    \includegraphics[width=0.19\textwidth]{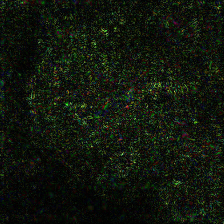}
     & 
     
    \hspace{-1em}
        \includegraphics[width=0.19\textwidth]{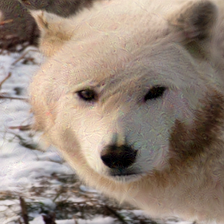}
        &
        \hspace{-1em}
        \includegraphics[width=0.19\textwidth]{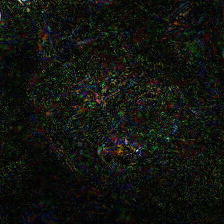}
     \\\includegraphics[width=0.19\textwidth]{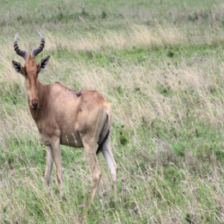}
     &
     \hspace{-1em}
    \includegraphics[width=0.19\textwidth]{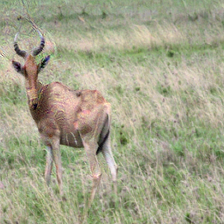}
     &
    \hspace{-1em}
    \includegraphics[width=0.19\textwidth]{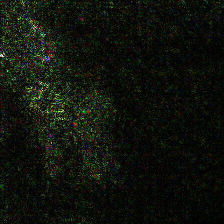}
     & 
     
    \hspace{-1em}
        \includegraphics[width=0.19\textwidth]{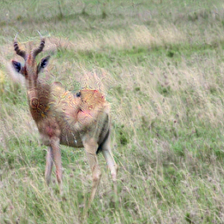}
        &
        \hspace{-1em}
        \includegraphics[width=0.19\textwidth]{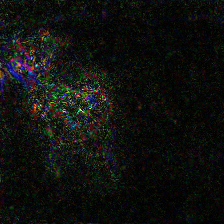}
     \\\includegraphics[width=0.19\textwidth]{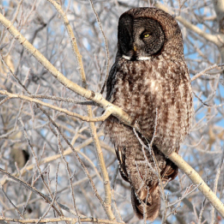}
     &
     \hspace{-1em}
    \includegraphics[width=0.19\textwidth]{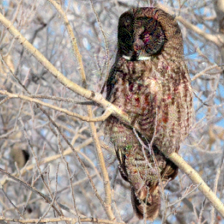}
     &
    \hspace{-1em}
    \includegraphics[width=0.19\textwidth]{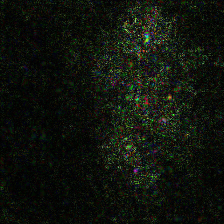}
     & 
     
    \hspace{-1em}
        \includegraphics[width=0.19\textwidth]{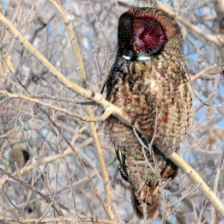}
        &
        \hspace{-1em}
        \includegraphics[width=0.19\textwidth]{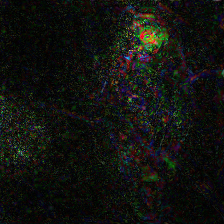}
     \\\includegraphics[width=0.19\textwidth]{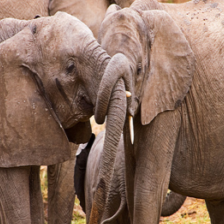}
     &
     \hspace{-1em}
    \includegraphics[width=0.19\textwidth]{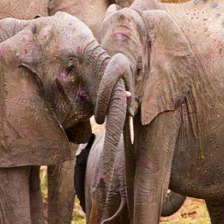}
     &
    \hspace{-1em}
    \includegraphics[width=0.19\textwidth]{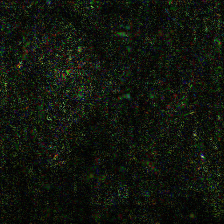}
     & 
     
    \hspace{-1em}
        \includegraphics[width=0.19\textwidth]{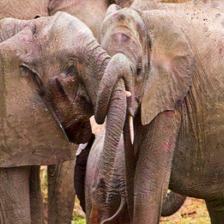}
        &
        \hspace{-1em}
        \includegraphics[width=0.19\textwidth]{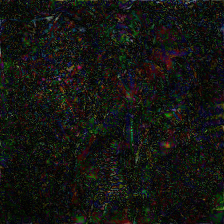}
     \\
     
    (a) & (b)  & (c)  & (d) & (e)
    
\end{tabular}}
\caption{Visualization of the experiment $\ell_2$-based adversarial attack with setting in~\ref{l2 setting}. (a) Original Image. (b) Adversarial image of DiffPGD. (c) Perturbation of DiffPGD. (d) Adversarial image of U-ScorePGD (Ours). (e) Perturabtion of U-ScorePGD (Ours). We scale up the perturbation's value by five times for better observation.
}
\label{fig:l2}
\end{figure*}

In Table~\ref{vit}, our method have almost 100\% attack success rate in white-box setting, and the best black-box attack success rate 60.4\%. In Table~\ref{vit with p}, our method, U-ScorePGD has the best anti-purify ability in the white-box setting and the second best anti-purify ability in the black-box setting, with ASR 78.9\% and 48.4\% respectively. Meanwhile, ScorePGD, which is also our method, has the best anti-purify ability in Swin-b, with 51.2\% attack success rate. We show the visualization of adversarial images of these experiments in Figure~\ref{fig:l2}. It is evident that the perturbations are concentrating on the main objective, similar to DiffPGD.

\begin{table}[ht]
\centering
\caption{Adversarial samples generated by PGD, DiffPGD and ours. The white-box scenario for surrogate models are represented with \colorbox{gray!20}{gray backgrounds}. The best result is in \textbf{bold}.}
    \begin{tabular}{c|cccc}
    \toprule
   Method & Vit-b-16 & Swin-b \\ 
   \midrule
     PGD   & \cellcolor{gray!20}{\textbf{99.9}\%}&  45.6\%  \\
    DiffPGD&   \cellcolor{gray!20}{98.5\%}&  57.7\%\\ 
    \midrule
    U-ScorePGD&  \cellcolor{gray!20}{99.6\%}&\textbf{60.4}\%\\
   \bottomrule
    \end{tabular}
    \label{vit}
    
\end{table}

\begin{table}[ht]
\centering
\caption{Adversarial samples generated by PGD, DiffPGD and ours against adversarial purification:
(+P) means the classifier is protected by an adversarial purification module. The white-box scenario for surrogate models are represented with \colorbox{gray!20}{gray backgrounds}. The best result is in \textbf{bold}, the second best is in \underline{underline} and the full knowledge setting regarding purification is denoted by \textcolor{lightgray}{light gray}.}
    \begin{tabular}{c|cc}
    \toprule
   (+P)Method & Vit-b-16 & Swin-b \\ 
   \midrule
     PGD   & \cellcolor{gray!20}{\underline{61.0}\%}&  19.2\%  \\
    DiffPGD&   \cellcolor{gray!20}{\textcolor{lightgray}{93.0\%}}&  34.4\%\\ 
    \midrule
    ScorePGD&  54.6\%&\textbf{51.2}\%\\
    U-ScorePGD&  \cellcolor{gray!20}{\textbf{78.9}\%}&\underline{48.4}\\
   \bottomrule
    \end{tabular}
    \label{vit with p}
    
\end{table}

\subsection{Results on MS COCO Object Detection}

Apart from classification, we would like to try our method on other task. In Table~\ref{obj detection}, we tested our method on 500 randomly selected images from the MS COCO validation dataset~\cite{lin2014microsoft}. We test FastRCNN-ResNet50~\cite{li2021benchmarking} for white-box attack and FastRCNN-MobileNet~\cite{ren2015faster} for black-box attack. Our method, U-ScorePGD, has the same attack performance as PGD, which brings down the clear mAP of FastRCNN-ResNet50 from 33.51\% to 0.02\% and has better transferability in FastRCNN-MobileNet, showing our method can be applied to other computer vision tasks. The attack setting for these experiments are $\gamma=16/255$, $\eta=2/255$ and $n=10$. Worth noting that we use the time-dependent classifier trained on ImageNet in resolution $512\times512$, since we consider the time-dependent classifier as a general image classifier that can be extended to the MS COCO dataset.

\begin{table}[ht]
\centering
\caption{mAP of clean samples and adversarial samples generated by PGD and ours. The white-box scenario for surrogate models are represented with \colorbox{gray!20}{gray backgrounds}. The best result is in \textbf{bold}.}
    \begin{tabular}{c|cc}
    \toprule
   Method & FastRCNN-ResNet & FastRCNN-MobileNet\\ 
   \midrule
   Clean &  33.51\% &   53.74\% \\
     PGD  & \cellcolor{gray!20}{\textbf{0.02}\%}& 45.03\% \\
    \midrule
    U-ScorePGD&  \cellcolor{gray!20}{\textbf{0.02}\%}&\textbf{43.31}\%\\
 
   \bottomrule
    \end{tabular}
    \label{obj detection}
    
\end{table}

\section{Limitation}

We are now choosing $t$ for the time-dependent classifier by trial and error. Selecting an adaptive $t$ for a specific image is yet an open question. In addition, our U-scorePGD sometimes gets stuck in a local maximum, just like multi-task training, due to the unbalanced optimization difficulty for the two losses. Finding a more advanced optimization method for our method is left to a future discussion.
One more limitation is that there are only a few open-source time-dependent classifier checkpoints, which are all trained on ImageNet, and hence, training a time-dependent classifier is needed for other datasets, especially private datasets.

\end{document}